# Testing-Driven Reliability Audit of Trajectory-Based Early Outcome Prediction for LLM Agents: Target-Specific Calibration Transfer Persists Within a Single Benchmark

**YanZe Cao**

Lizhi College, Shaanxi University of Technology, No. 1 East 1st Ring Road, Hantai District, Hanzhong, Shaanxi 723001, China

Email: caoyanze426@gmail.com

ORCID: 0009-0000-1332-2074

## *Abstract*

Predicting early outcomes based on trajectory can decrease the expenses associated with agent evaluation by terminating a run once the outcome becomes sufficiently predictable, assuming that the predictor's confidence is properly calibrated. Calibration is at risk when a predictor is applied to an agent on which it was never trained, but it is not known whether such transfer failures are broad across agent systems or concentrated in specific target agent/head combinations. Using public SWE-bench Verified trajectories and a frozen dual-head early-outcome prediction pipeline, we ran a leave-one-agent-out calibration audit, a shared-predictor leave-two-agents-out control, oracle prior correction, and a robustness battery over training cohorts, task resampling, task halves, jackknife, and thresholds. Fixed-scaffold TerminalBench analysis served as a pre-registered boundary test. Broad same-predictor pairwise heterogeneity was not supported; the median pairwise corrected-gap differences were 0.0180 (SUCCESS head, 45 pairs) and 0.0385 (FAILURE head, 35 pairs), and the pre-registered heterogeneity criterion was not met on either head. Two specific combinations, gpt-5-mini/SUCCESS and claude-opus-4.6/FAILURE, showed persistent calibration-transfer errors (median corrected gaps 0.1377 and 0.1107) without a sign reversal under any frozen control. TerminalBench did not establish cross-benchmark replication: the success target produced zero decisions (INDETERMINATE), and the failure target did not satisfy the pre-registered persistence criterion. Therefore, a strong target-specific calibration-transfer error can exist within one frozen environment, but the evidence does not establish that the error is intrinsic to the model or general across benchmarks.

## *1. Introduction*

Agentic coding runs are long, and evaluating them step-by-step is expensive: every step a coding agent takes is paid for during benchmarking and deployment. If a run's eventual success or failure is already predictable from its prefix, that cost can be reduced by stopping early, and early outcome prediction has been introduced for this purpose, together with calibrated stopping policies and transfers to unseen agents [1]. The measurement problem behind this line of work is well documented: agent benchmarks are costly, and accuracy-only reporting hides cost and variance [2], while reports on early aborts and early uncertainty in long-horizon agents argue from different angles that the value of stopping early is worth studying [3,4]. Throughout this paper, early stopping means stopping an agent running at a trajectory prefix; it does not mean stopping a reasoning chain, which is a different object studied elsewhere [5,6].

An early stop at high confidence is only safe if that confidence is calibrated; a stop taken because success appears highly likely must correspond to a realized success rate close to the stated confidence. Calibration has a long methodological history [7], and calibration validity is known to deteriorate under a distribution shift, including shifts between domains and between models [8,9,10]. In the study of agent systems, calibration research has primarily focused on how confident an agent is regarding its own actions. This issue has been specifically defined as agentic confidence calibration [11]. The object of calibration in this study is different: we audit the confidence of an externally trained trajectory-based outcome predictor that is applied to an agent it was not trained on.

Aggregate evaluation can hide behavior at the level of individual targets, a phenomenon that is itself established; global calibration comparisons can reverse under accuracy control [12]; pooled monitor accuracy can conceal fragility on a subset [13]; and averaging trajectories together can hide stage-specific overconfidence that step-level calibration does not reveal [14]. The reporting of early outcome predictions has followed a similar aggregate pattern, summarizing per-agent transfer behavior as an average perturbation [1]. What is not known from existing reports is whether cross-agent calibration transfer failures are broad across agent systems or concentrated in specific target agent/head combinations. The distinction has practical weight: if the transfer error is broad, a fixed confidence threshold is generally unsafe for new agents, whereas localized transfer error would instead motivate target-specific checks before deployment.

This study addressed this question in a frozen evaluation environment. We adopted the reference-free dual-head early-outcome predictor design, calibrated stopping policy, and leave-one-agent-out transfer protocol introduced in prior work [1]. We kept the predictor configuration and stopping-policy design fixed across the audited analyses. We added a shared-predictor control that holds one trained predictor and calibrator fixed across two unseen agents, an oracle prior correction that is an established label-shift adaptation used here only as a diagnostic [10], a persistence battery over training cohorts, task resampling, task halves, partner-fold jackknife, and stopping thresholds, and a fixed-scaffold boundary test on a second benchmark [15]. The manuscript and the subsequent audit stages consumed the frozen experimental record, and no new predictions were generated during manuscript preparation. Figure 1 summarizes the design and frozen evidence ladder.

The results can be divided into two parts. First, broad same-predictor pairwise heterogeneity was not supported: under shared-predictor control, the median pairwise corrected-gap difference was 0.0180 on the SUCCESS head across 45 eligible pairs and 0.0385 on the FAILURE head across 35 eligible pairs, and the pre-registered heterogeneity criterion was not met on either head (SAME_PREDICTOR_HETEROGENEITY = FALSE). Second, two frozen target/head combinations were persistent on SWE-bench: gpt-5-mini/SUCCESS (median corrected gap 0.1377 across nine eligible occurrences) and claude-opus-4.6/FAILURE (median 0.1107 across eight eligible occurrences), both retaining sign under task-cluster bootstrap, oracle prior correction, task halves, partner-fold jackknife, and stopping thresholds. A fixed-scaffold TerminalBench analysis did not establish a cross-benchmark replication of this pattern (Phase 2B: NO_CROSS_BENCHMARK_SIGNAL; Phase 2B-D: NO_CROSS_BENCHMARK_REPLICATION; Phase 2E: INSUFFICIENT_CROSS_BENCHMARK_SUPPORT).

This study makes three contributions. First, a target-level audit of an existing reference-free early outcome predictor: rather than characterizing cross-agent calibration transfer in general, we report signed per-target/head calibration gaps under leave-one-agent-out transfer and test whether the predictor's transfer error is broad across agents or concentrated in specific target/head cells, with the broad-heterogeneity gate not met and the residual concentrated. Second, we present a leave-two-agents-out shared-predictor control that fixes the trained predictor and its calibrator inside each pair so that the difference between two unseen targets cannot be attributed to retraining, together with the finding that the two named target/head errors keep sign and magnitude across changing source cohorts, task resampling, partner jackknife, task halves, and stopping thresholds, with prior correction credited as an existing label-shift adaptation used here only as a diagnostic. Third, we present a pre-registered exact-identity boundary test in which the two frozen SWE target cells are re-evaluated on TerminalBench under a fixed scaffold; the result is a negative and partly indeterminate boundary that constrains how the SWE phenotype can be read, rather than a claim that benchmarks differ in general. The predictor architecture and transfer protocol are not claimed as contributions; they are reused from prior work [1], and the adjacent agentic-calibration framing belongs to its own line of work [11].

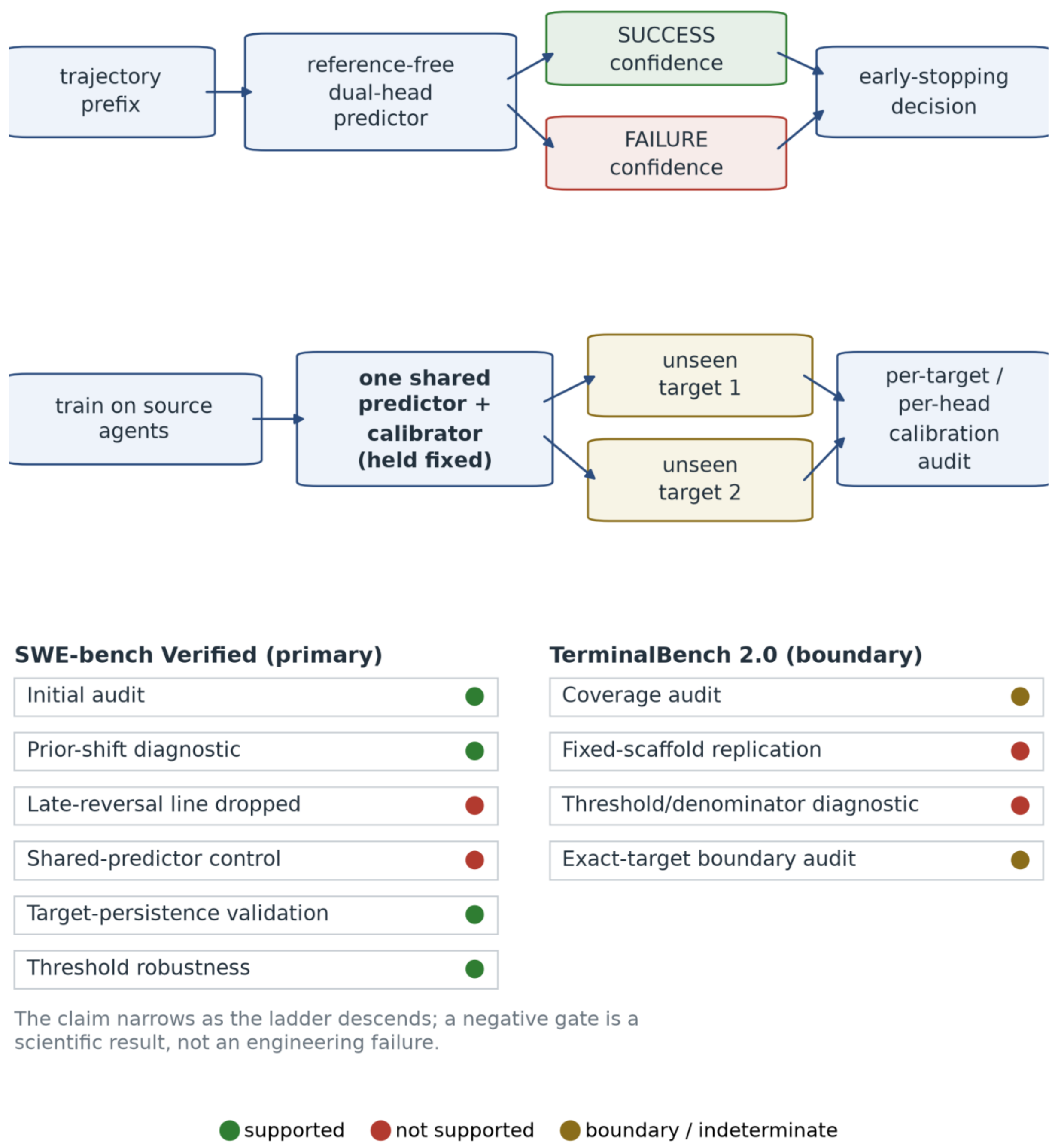

Figure 1. Study design and evidence flow for trajectory-based early outcome prediction under cross-agent transfer. Panel A: A trajectory prefix is scored by a frozen reference-free dual-head predictor into SUCCESS/FAILURE confidence, in which the frozen safe-stop policy turns into an early stopping decision (success_thr = failure_thr = 0.95, min_step = 0, consecutive = 1). Panel B: The predictor is trained on source agents and applied unchanged to a held-out unseen target agent, whose per-head calibration is then audited. Panel C: Evidence ladder, listing the SWE-bench Verified phases (0D initial audit, 0E oracle prior-shift decomposition, 1A same-predictor leave-two-agent-out control, 1A-D target persistence validation, 1B threshold robustness) and the TerminalBench phases (2A feasibility, 2B fixed-scaffold replication, 2B-D

denominator diagnostic, 2E exact-target boundary audit) with their frozen gate outcomes, listed here as frozen: Phase 0D GO_CANDIDATE; Phase 0E GO; Phase 0C SIGNAL_NOT_YET_ROBUST; Phase 1A SAME_PREDICTOR_HETEROGENEITY = FALSE (PARTIAL_PASS); Phase 1A-D STRONG_TARGET_SPECIFIC_SIGNAL; Phase 1B STRONG_PASS; Phase 2A measured design selection (no gate); Phase 2B NO_CROSS_BENCHMARK_SIGNAL; Phase 2B-D NO_CROSS_BENCHMARK_REPLICATION; Phase 2E INSUFFICIENT_CROSS_BENCHMARK_SUPPORT. Unit of analysis: Study stage (not an individual trajectory). No numerical results were plotted. The ladder is shown to make explicit that the study progressively narrows its claim: stages with negative gate outcomes are part of the design, and a negative gate is not an engineering failure. Panel C does not establish any magnitude, and no panel in this figure reports a measured effect.

## *2. Related Work*

### 2.1 Early Agent Outcome Prediction and Efficient Evaluation

Early outcome prediction from a trajectory prefix was introduced using a reference-free dual-head predictor, calibrated decision thresholds, a leave-one-agent-out evaluation protocol, and transfer splits to unseen models and scaffolds [1]. That work establishes the setting adopted here, including the policy that converts head confidence into a stop decision, and reports aggregate transfer behavior across its splits. The motivation for cheaper evaluation is broader than any single method: agent benchmarking is expensive relative to the decisions it supports, and accuracy-only reporting obscures costs [2]. A second and mostly separate efficiency axis reduces the number of benchmark tasks rather than the length of a run, for example, by evaluating language models with fewer examples [16]; work in that family is not comparable to within-trajectory stopping and is not the object of this study. Adjacent systems abort doomed episodes with recall-controlled probes [3], and one report argues that early uncertainty in long-horizon agents is informative, mainly near completion [4]. A further family stops reasoning traces rather than agent runs [5,6]; these methods act on a different object and are cited here only to keep the terminology distinct.

How this study differs: We neither reduced the task set nor shortened the reasoning chains, and we did not introduce a stopping method. We took the existing predictor and protocol and asked where the predictor's confidence went wrong when the agent changed at the granularity of the target/head cells.

### 2.2 Confidence Calibration for LLMs and Agents

Post-hoc calibration of neural network confidence is a mature methodology, with temperature scaling as the canonical example and the expected calibration error as a standard summary [7]. For agentic systems, the problem of agentic confidence calibration has been explicitly introduced, reporting that agents are overconfident and that calibration must be studied at the trajectory level [11]. Calibration under shift is also an active area: conformal methods with test-time training aim to provide generalizable validity across domains [8] and further line models confidence before and after reasoning and reports degradation out of domain [9]. The label-shift literature establishes that the relationship between predicted confidence and realized outcome rate changes when class priors change and provides correction machinery for it [10]. One recent report found that early uncertainty carries predictive value only near completion, an example of stage-dependent calibration behavior [4].

How this study differs: The works above calibrate what a model or an agent says about itself or recalibrate a model against new domains. This study audits a different object, the confidence of an externally trained trajectory-based outcome predictor, and asks whether that confidence stays calibrated when the predictor is applied to a new agent system. No recalibration method was proposed or evaluated in this study.

## 2.3 Trajectory-Level Agent Evaluation

Trajectory-level evaluation has developed into a distinct line of work. TIDE decomposes diagnostic information along trajectories [17], ToolPRMBench evaluates process reward models for tool-using agents [18], trajectory-judge shows what outcome-only judges miss on agent trajectories [19], and Counsel provides a meta-evaluation dataset for judge critiques on agentic tasks [20]. These works score, diagnose, or meta-evaluate completed trajectories and their judges.

How this study differs: none of them predicts a final outcome from a prefix, none reports the calibration of a stopping decision, and none uses a held-out-agent design. Our analysis intervenes earlier in the trajectory lifetime, when a policy would decide to stop, and reports the reliability at that point.

## 2.4 Calibration Transfer Across Agent Systems

Calibration conclusions are unstable across models and benchmarks. An accuracy-controlled comparison shows that calibration rankings can reverse once accuracy is held fixed [12], a pooled monitor metric can be dominated by an easy subset and conceal fragility on the remainder [13], and averaging trajectories together can hide stage-specific overconfidence even when step-level calibration looks acceptable [14]. Two further lines push in the same direction: experiential confidence estimation across reasoning and agent settings [21] and calibration of agent confidence from internal representations [22]. Early outcome prediction reports transfer behavior in aggregate form, summarized as an average perturbation per agent [1].

How this paper differs: The existence of aggregation artifacts and benchmark dependence is taken here as established. The narrower question is whether a predictor's transfer error is broad across held-out agents or concentrated in specific target/head cells. This question is answered with a fixed-predictor pair control rather than a new calibration method. The frozen record returns a negative broad gate, together with a positive localized residual, and a second-benchmark boundary test that does not establish the phenotype beyond the first benchmark.

## *3. Methods*

The methods described below describe only procedures that were executed in the frozen record, and the configuration quantities restated were taken from the frozen artifacts recorded in the supplement.

## 3.1 Problem Setting

A run is a trajectory $\tau = (s_1, \ldots, s_T)$ of an agent solving one task with a binary terminal outcome $y$ (the task is resolved or not). The prefix $\tau_{1:t}$ represents the first $t$ steps of that trajectory. Early outcome prediction trains two heads on prefix features: a SUCCESS head that estimates the probability that the run ends in a resolved outcome and a FAILURE head that estimates the probability that it does not. A stopping policy converts the head's confidence at a prefix into a decision to stop, and the confidence attached to that decision is the object of the audit.

The unit of transfer is the target agent or agent system: the model (and, on the second benchmark, the scaffold), whose trajectories are held out and scored by a predictor trained on the remaining source-agent cohort. Within one target, each prediction head is audited separately, and the pair (target agent and head) is called a target/head cell.

The stopping policy that converts head confidence into a decision is dual: at each prefix, the two heads are scored, and a decision is made at the first prefix at which a head's calibrated score crosses its threshold, with the minimum step set to zero and the consecutive-crossing requirement set to one. The score attached to that decision is the confidence whose calibration is audited, and the decision step is recorded so that a trajectory-level outcome can be compared with the confidence expressed about it. Because the policy is deterministic, the same frozen configuration produces identical decisions when it is re-applied to frozen per-prefix predictions, which is what the threshold sweep of Section 3.8 exploits.

Throughout, the signed calibration gap of a cell is the mean calibrated decision score at the stop point minus the realized outcome rate (empirical precision) of the decided rows in that cell, exactly as frozen in the audit protocols. A positive SUCCESS-head gap, therefore, means that success confidence at the stop point exceeds the observed success rate; the same convention, predicted minus realized, applies to the FAILURE head. All gap values reported in the results are signed, and intervals and ranges are computed on the frozen point values.

### 3.2 Datasets and Trajectory Representation

The primary corpus is a frozen public collection of SWE-bench Verified agent trajectories (source revision `tarsur385/swebench-verified-trajectories@773748a7c1222e8a642a7059821498e14293562a`). It contains 5,000 raw trajectories, of which the frozen adapter accepts 4,989 and rejects 11, covering 10 model labels and 500 shared tasks. The Phase 0B signal hunt scored 235,968 prefix rows on the held-out folds of the leave-one-model-out design. A trajectory is the sequence of steps of one model on one task instance; the prefix representation, feature construction, and adapter behavior are consumed as frozen and are not modified here.

The second corpus was a community-compiled TerminalBench trajectory dataset (revision `04e8940f5b6736a7ce8d22224fe2f2af74163ed2`, license apache-2.0). At the dataset-audit level, it contains 52,104 public rows, of which 34,462 carry a usable trajectory (17,642 rows have no trajectory steps), covering 89 tasks across 49 model labels (33 usable) and 26 scaffolds (9 usable), that is, 109 model–scaffold combinations of which 70 are usable. The fixed-scaffold replication analysis (Phase 2B) runs inside the `terminus-2` scaffold and uses 15,889 usable trajectories across 89 tasks after the frozen step-builder drops, grouped into 29 eligible exact models. The frozen analysis universe scored 424,413 prefix rows with 5,178 successes and 10,711 failure trajectories.

The two corpora were not treated as independent samples. The TerminalBench trajectory source is community-compiled, and a partial provenance overlap can be verified between that community-compiled dataset and the historical TerminalBench corpus/provenance used by the upstream EarlyEval release; the trial-level identity overlap between the community-compiled dataset and that historical corpus/provenance could not be verified from the frozen record, and no exact overlap percentage is inferred. TerminalBench was therefore used as a boundary test of a fixed-scaffold evaluation setting, not as an independent replication dataset in a strict trial-level sense, and its three trajectory counts (52,104 dataset rows, 34,462 usable rows, and 15,889 analysis-universe trajectories) were kept separate throughout (Table 1).

| | | |
|---|---|---|
| Trajectory source | tarsur385/swebench-verified-trajectories @ 773748a7c1222e8a642a7059821498e14293562a | community-compiled TerminalBench 2.0 trajectories @ 04e8940f5b6736a7ce8d22224fe2f2af74163ed2 (apache-2.0) |
| Usable trajectories | 4,989 adapter-PASS (11 adapter-FAIL) out of 5,000 raw trajectories | 52,104 public rows; 34,462 rows with usable trajectory; 15,889 trajectories entering the frozen Phase 2B analysis after step-builder drops |
| Tasks | 500 shared tasks across all 10 model labels | 89 tasks |
| Target-system definition | 10 frozen model labels (agent), each paired with the SUCCESS and FAILURE prediction head | 49 models (33 usable) x 26 scaffolds (9 usable) = 109 combos (70 usable); frozen Phase 2B universe = 29 terminus-2 models |
| Primary split / control | leave-one-model-out; plus leave-two-agents-out shared-predictor pair folds (Phase 1A) | fixed-scaffold (terminus-2) leave-one-model-out cross-benchmark replication; exact-model identity audit |
| Primary uncertainty unit | held-out model / agent, with task-cluster bootstrap over instance_id | held-out model within the fixed scaffold; task-cluster bootstrap |
| Role in paper | primary benchmark; all positive continuous results | second-benchmark boundary test; no replication claim |

Footnotes. (1) TerminalBench trajectory counts are reported at three distinct scopes and are labelled in-cell: 52,104 public dataset rows, 34,462 rows with a usable trajectory, and 15,889 trajectories entering the frozen Phase 2B analysis universe after step-builder drops; they must not be collapsed into a single count. (2) The frozen Phase 2B universe comprises 29 terminus-2 models drawn from 49 models (33 usable) and 26 scaffolds (9 usable) across 109 model-scaffold combinations (70 usable). (3) Unit of analysis: benchmark-level design description, not a measured outcome.

Table 1. Datasets and experimental design. Two corpora are used: SWE-bench Verified (primary) and TerminalBench 2.0 (second-benchmark boundary test). For each corpus, the table records the trajectory source and revision, the usable trajectory counts, the task count, how the target system is defined, the primary split and control, the primary uncertainty unit, and the role in the paper. SWE-bench Verified counts were reported at the level of the frozen corpus (4,989 adapter-PASS trajectories of 5,000 raw, 11 adapter-FAIL) over 500 shared tasks and 10 frozen model labels. TerminalBench counts are reported at three distinct scopes and are labelled as such: 52,104 public dataset rows, 34,462 rows with a usable trajectory, and 15,889 trajectories entering the frozen Phase 2B analysis universe after step-builder drops; the frozen Phase 2B universe is 29 terminus-2 models drawn from 49 models (33 usable) and 26 scaffolds (9 usable) across 109 model-scaffold combinations (70 usable). Unit of analysis: benchmark-level design description, not a measured outcome. The table supports the definition of the analysis universes and does not report or imply any replication results.

### 3.3 Early Outcome Predictor

The predictor audited here is the frozen reference-free dual-head predictor pipeline carried through the study as `P0B_ReferenceFree_LightGBM_Dense_AF_FoldLocalFE`. It follows the early outcome-prediction design: behavioral and textual prefix features feed two LightGBM heads, one per outcome direction, whose calibrated scores are thresholded by a dual-stopping policy. In the frozen configuration, the feature family actually used is the action-feedback family (TF-IDF over action and feedback text in five blocks), with the whole gold-answer family removed so that no reference solution leaks into the features. The feature engineer is fitted on the training split only, so the held-out agent contributes no fitted parameter, including the TF-IDF vocabulary, idf, SVD basis, numeric scaler, label encoders, and calibration and threshold selection. The frozen stopping policy was dual, with `success_thr = failure_thr = 0.95`, `min_step = 0`, and `consecutive = 1`. Trajectories in which neither head crosses its threshold produce no decision and are recorded as no-stop. Exact frozen hyperparameters: The two heads were LightGBM binary classifiers (`objective = binary`, `metric = binary_logloss`) trained with `learning_rate = 0.05`, `num_leaves = 63`, `max_depth = 8`, `min_child_samples = 50`, `subsample = 0.8`, `colsample_bytree = 0.8`, `reg_alpha = 0.1`, `reg_lambda = 1.0`, and `seed = 42`. Decision scores were calibrated using a validation-only Platt sigmoid fitted on the validation logits. The per-step feature vector has 435 columns: 115 dense action and feedback features plus five TF-IDF/SVD text blocks of 64 components each (task prompt, prefix action, prefix feedback, last action, last feedback), with 1-2 grams, minimum document frequency 5, at most 30,000 terms per block, sublinear term frequency, SVD random seed 42, and a dense-feature standardization step. The complete 115-name dense-feature list, frozen per-fold matrix shapes, and cross-phase parity check are given in the supplement (Section S9.3).

The predictor architecture was not a contribution of this study. It was introduced in prior work on early outcome prediction [1], and the present study holds the same frozen configuration fixed in every phase, including the second benchmark.

### 3.4 Cross-Agent Calibration Audit (Phase 0D)

The initial audit holds one model label at a time across the 10 frozen labels. For each held-out model, the predictor trained on the remaining nine models produced calibrated prefix scores, and the frozen stopping policy was applied to those scores. For each (held-out model, head) cell, the audit reports the number of decisions, the empirical precision of the decided rows, the mean calibrated decision score at the stop point, and their difference, and the signed calibration gap. Uncertainty on cell-level and range statistics is obtained by a task-cluster bootstrap that resamples task identifiers with replacement, 2,000 replicates, and seed 42, with replicates in which a statistic is undefined dropped and counted.

Three qualification signals were pre-registered for this stage: a precision range across models of at least 0.15, with a bootstrap lower bound above 0.05, in either head; an association between the held-out outcome prior and the error-rate-based calibration behavior in the expected directions; and at least three models with at least 20 decisions and an absolute calibration gap of at least 0.10, in at least one head. This stage is explicitly a qualification step for later controlled tests; the pairwise heterogeneity question is decided by the shared-predictor control of Section 3.6. One held-out model (gemini-3-pro) has a degenerate SUCCESS-head prior on its test folds; its FAILURE-head cell is marked as prior-degenerate wherever it appears, and its values are excluded from the residual gate of Section 3.5 rather than being smoothly imputed.

The two bookkeeping rules bound the denominators. Decisions are attributed to a (model, head) cell only when the policy actually stops the trajectory at that head; trajectories that never cross a threshold are no-stop rows and enter no calibration denominator, and missing denominators are recorded as NA rather than being imputed. The corpus-level pooled row produced by the audit is a reporting aggregate over the same decided trajectories and is excluded from every per-model count and gate.

### 3.5 Oracle Prior-Shift Diagnostic (Phase 0E)

The second stage asks whether the audit's cross-agent pattern survives the target's own outcome prior as an explanation. For each held-out model, the frozen diagnostic computes the model's training and test success priors and forms the odds-ratio coefficient $r_m = [\pi_{test} / (1 - \pi_{test})] / [\pi_{train} / (1 - \pi_{train})]$. Every stop-point head probability is then shifted in log-odds space by $\ln r_m$ for the SUCCESS head and by exactly $-\ln r_m$ for the complementary FAILURE head, after which the calibration gaps are recomputed. The coefficient is an oracle quantity computed from the realized labels of the held-out model. The shift is analytic, does not change any early decisions, and adjusts the interpretation of confidence after the decision has been made. Correction is an established label-shift adaptation, not a new method, and is used here only as a diagnostic method [10]. The stage reports pooled raw and corrected gaps, the residual gap ranges per head with task-cluster interval estimates, and a residual gate that counts non-degenerate models with at least 20 decisions and an absolute corrected gap of at least 0.10 in at least one head. The correction removes one compositional explanation of the pattern; it does not establish causal independence from the prior outcome, and it is not reported as doing so.

Input identity was verified before the correction was interpreted: the recomputed raw scores, precisions, gaps, and priors reproduce the Phase 0D values within $1.11 \times 10^{-16}$ (declared tolerance $1 \times 10^{-9}$).

### 3.6 Same-Predictor Leave-Two-Agent-Out Control (Phase 1A)

The central control removes the alternative explanation that the two target agents differ merely because they are scored by differently trained predictors. The design forms all 45 unordered pairs of the 10 model labels, leaves both members of a pair out of training, trains one shared predictor and calibrator on the remaining eight agents, and applies that single model to both held-out targets. For each pair and head, the control computes the corrected calibration gap of each target on the pair's common task support and takes the pairwise difference between the two targets' gaps. The pair-level statistic is the median of the differences across pairs, with a task-cluster bootstrap (2,000 valid replicates, seeds 4201 and 4202 for the two heads).

Two pre-registered criteria were evaluated per head: at least 20 eligible pair folds, where a pair fold is eligible for a head when both held-out members carry at least 20 early decisions in that head, and a heterogeneity criterion requiring a median pairwise difference of at least 0.05, with a bootstrap lower bound above 0.03. A head was flagged as showing same-predictor heterogeneity only if every condition of its own rule held; the frozen result was recorded per head, and the overall same-predictor gate was recorded as satisfied when either head passed, that is, as a disjunction rather than a conjunction. Independent of the pair gate, the stage scans all 20 target/head cells for persistence using a pre-registered rule (at least seven eligible occurrences, a median absolute corrected gap of at least 0.08, and a same-sign fraction of at least 7/9, with an 80 percent alternative when exactly seven or eight occurrences are eligible). All 45 pair folds were planned, completed, and retained, with zero fold failures.

Each pair's statistics are computed on the task support shared by its two held-out members, and the pair's priors are measured on the same support; the full-universe priors are not used for the gate. Occurrences in which a held-out target has fewer than 20 decisions or a degenerate prior are excluded from the persistence scan rather than imputed, and exclusion counts are reported per cell.

### 3.7 Target-Specific Persistence Tests (Phase 1A-D)

The two cells that satisfied the Phase 1A persistence rule were frozen as primary targets before the robustness analysis was run: gpt-5-mini/SUCCESS as PRIMARY_TARGET_1 and claude-opus-4.6/FAILURE as PRIMARY_TARGET_2. A third cell, minimax-m2.5-high/FAILURE, was declared sensitivity-only and was never promoted to primary evidence. The analysis reuses the same-predictor occurrences of each target across partner folds and evaluates four pre-registered criteria on the frozen point values: a minimum of seven eligible occurrences, a median absolute corrected gap of at least 0.08, a same-sign fraction of at least 0.8, and a task cluster bootstrap interval (5,000 replicates, seed 42031) that excludes zero while retaining the baseline sign. Two further controls are required: a partner-fold jackknife in which the signed median must keep its sign with a minimum absolute value of 0.06 in every leave-one-partner-out refit, and a deterministic task-half split in which each half must contain at least five eligible folds and keep the baseline sign with an absolute magnitude of at least 0.06; a fold counts as eligible in half when that half carries at least 10 early decisions. The thresholds and selection rules were frozen before this robustness analysis; no target was reselected after seeing these results.

### 3.8 Threshold Robustness (Phase 1B)

The stopping threshold is a policy parameter; therefore, the persistence of each primary cell is re-evaluated on the frozen grid 0.900, 0.925, 0.950, and 0.975 by re-applying the unchanged policy to the same frozen per-prefix predictions, with per-threshold task-cluster bootstrap intervals (2,000 replicates, fixed seeds). The 0.950 anchor was reproduced against the frozen decision table before any sweep value was interpreted: all 44,901 frozen decision rows agreed on the decided flag, decision, decision step, and decision score, with a maximum absolute difference of $1.11 \times 10^{-16}$, that is, one floating-point unit at the declared tolerance. A threshold point is counted as robust for a target when it has at least seven eligible occurrences, a median absolute corrected gap of at least 0.06, a same-sign fraction of at least 0.75, and a task-cluster bootstrap 95% confidence interval for the median signed gap that excludes zero and retains the baseline Phase 1A-D sign; a target's sweep passes when at least three of the four points pass and the reference threshold 0.950 passes. Where an occurrence universe does not clear the decision minimum, the point is recorded as not evaluable, and no value is imputed. No predictor is retrained, and no calibrator is refitted in this stage.

### 3.9 TerminalBench Boundary Study

The boundary study proceeded in four frozen stages. Phase 2A audits the coverage of the community-compiled TerminalBench trajectory dataset and selects a feasible fixed-scaffold design. With the scaffold held at `terminus-2`, the exact underlying model is the only varying dimension that keeps the scaffold identity out of the comparison, while allowing a leave-one-model-out replication of the audit on a second benchmark.

The design selection was measured rather than assumed. From the audited coverage of 52,104 dataset rows, the scaffold-controlled candidate with a single shared scaffold carried 29 eligible model targets and 406 model pairs, with common task support between 86 and 89 tasks per pair (median 89) and a mean usable fraction of 0.985858 per target; this candidate became the Phase 2B universe.

Phase 2B runs replication with 29 eligible models: each fold holds out one exact model, trains on the remaining 28, and scores the trajectories of the held-out model under the same dual policy at the primary threshold of 0.950. A target/head cell is flagged as a large-gap target when it has at least 20 decisions, an absolute corrected gap of at least 0.08, and a task-cluster bootstrap interval that excludes zero while retaining the sign. The cross-benchmark gate requires at least three such models in at least one head, at least two of them with an absolute gap of at least 0.10, and not all of those models from a single provider family. Phase 2B-D is a threshold and denominator diagnostic, not a replacement for the primary gate; it repeats the large-gap classification at 0.900 and 0.925 alongside the 0.950 anchor, reports the eligibility counts per head and threshold, and introduces one feasibility requirement of its own, HEAD_DENOMINATOR_ADEQUATE, which holds when a head carries at least eight eligible targets; that requirement belongs to the diagnostic and is not part of the Phase 2B primary gate.

Phase 2E revisits the two SWE primary targets at the exact model identity. Identity is taken from literal dataset metadata, never from name similarity: the SWE label `gpt-5-mini` corresponds to `gpt-5-mini@openai` on TerminalBench, and `claude-opus-4.6` corresponds to `claude-opus-4-6@anthropic`. For each target, the stage applies the pre-registered four-clause status rule at the primary threshold 0.950 (persistent, collapsed, sign-changed, or indeterminate) and reports the resulting classification, including cases that the rule does not cover, which are recorded literally and treated as not persistent.

### 3.10 Statistical and Reproducibility Protocol

All statistics reported in this paper are frozen: task-cluster bootstrap intervals, Spearman correlations, medians, ranges, decision denominators, and gate outcomes. No p-value, effect size, Bayesian quantity, meta-analytic combination, or new interval was computed for this manuscript, and no threshold was changed after a result was seen. Gate definitions and their constants are quoted from frozen protocol files, and each phase records its own execution and integrity checks.

Three verification steps bound the numbers used. First, every phase verifies the hashes of its inputs against the manifest of the previous phase before computing anything. Second, where a phase re-derives earlier decisions, it must reproduce them exactly: the 0.950 anchor of the threshold sweep reproduces all 44,901 decision rows (maximum absolute difference $1.11 \times 10^{-16}$), and the TerminalBench 0.950 anchor reproduces all 15,889 decision rows with 3,784 decided and zero mismatches. Third, the prior-correction stage reproduces the audit's raw scores, precisions, gaps, and priors to within $1.11 \times 10^{-16}$. All executions were offline with respect to the model APIs: the phases consumed frozen trajectories and predicted locally, and the phase records reported zero API calls and zero LLM calls.

## *4. Results*

### 4.1 Initial Cross-Agent Calibration Audit

The audit consumes the frozen decision tables of the signal-hunt stage, which applies the dual policy to 235,968 held-out prefix rows and produces 2,822 early decisions: 2,333 correct SUCCESS decisions, 305 correct FAILURE decisions, 132 false-success decisions, 52 false-failure decisions, and 2,167 no-stop trajectories. Only the decided trajectories enter a cell's calibration denominator; no-stop trajectories enter none. All 10 model folds have a defined decision denominator in both heads. At the stop point, SUCCESS-head precision ranges from 0.7830 (gpt-5-mini, 235 decisions) to 1.0 (gemini-3-pro, 258 decisions), a range of 0.2170 with a task-cluster bootstrap interval of [0.1660, 0.2709]. The FAILURE-head range is 1.0 (gemini-3-flash-high at precision 1.0; gemini-3-pro at 0/25, precision 0.0), but that qualification is a degenerate-fold artifact: excluding the prior-degenerate gemini-3-pro fold, the FAILURE-head range falls to 0.1475 with interval [0.1094, 0.2727] and no longer clears the 0.15 range criterion. The SUCCESS head qualifies independently.

Pooled across the model folds, the mean calibrated decision score exceeded the realized outcome rate by 0.0194 on the SUCCESS head (mean score 0.9659) and by 0.1063 on the FAILURE head (mean score 0.9607). Four target/head cells met the raw decision-count and gap thresholds (at least 20 decisions and an absolute gap of at least 0.10): gpt-5-mini/SUCCESS (gap 0.1864 over 235 decisions), claude-4.5-opus-high/FAILURE (0.1149 over 61 decisions), claude-opus-4.6/FAILURE (0.1110 over 41 decisions), and the prior-degenerate gemini-3-pro/FAILURE cell (0.9582 over 25 decisions at precision 0.0). The prior-degenerate cell was excluded from gate counting, leaving three non-degenerate cells contributing to the qualification. The held-out success prior is associated with the SUCCESS-head error in the expected direction (Spearman rho = −0.7915, interval [−0.9605, −0.5897]) and FAILURE-head error (rho = 0.6442, interval [0.2917, 0.8283]); both associations retain sign after excluding gemini-3-pro (rho = −0.7120 and 0.5085). All three qualification signals are true, and the stage records GO_CANDIDATE, which is a screening outcome that motivates the controlled tests below, not evidence of broad heterogeneity.

This stage is the point where an earlier analysis direction was dropped: the pre-registered late-reversal analysis returned SIGNAL_NOT_YET_ROBUST, and the program redirected to the cross-agent reliability question reported here. This negative intermediate result was retained in the analysis record.

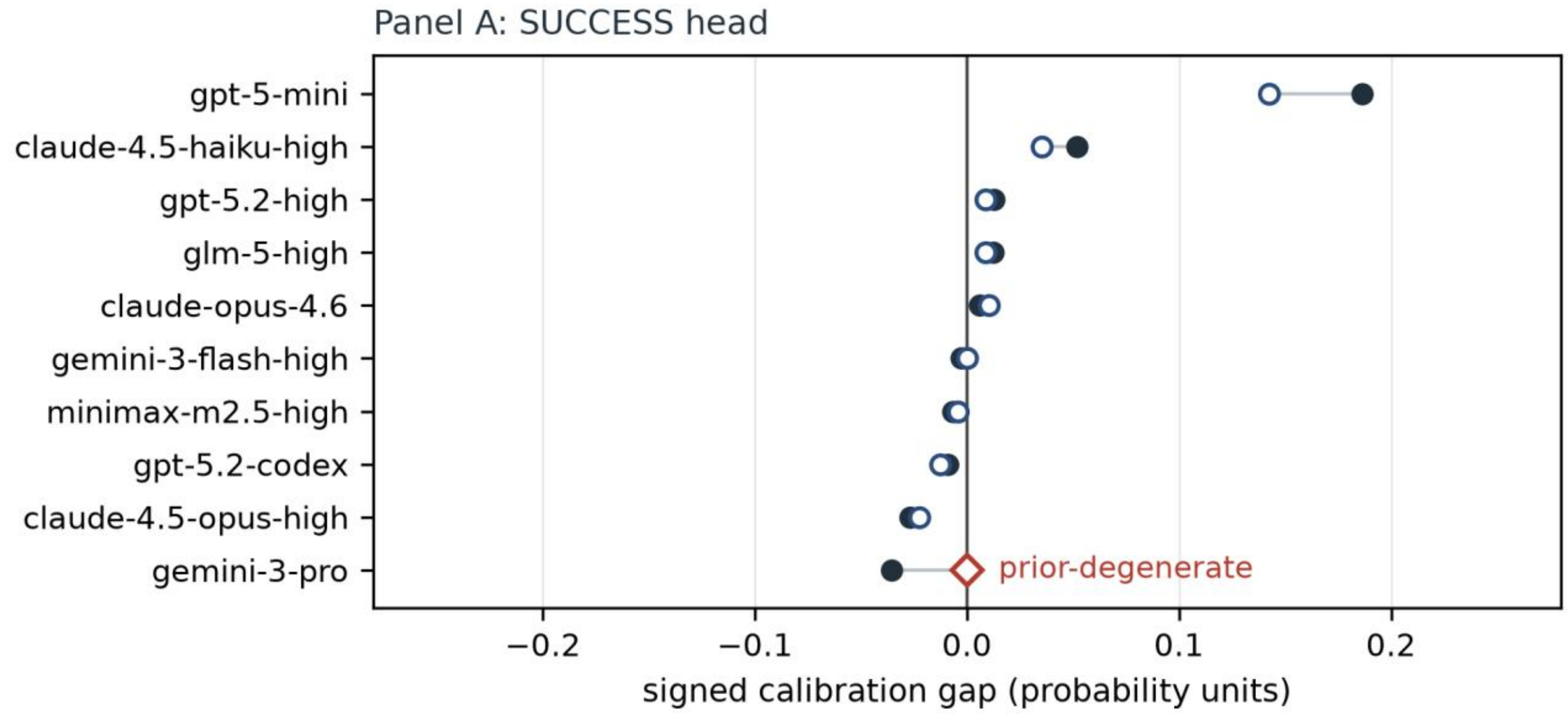

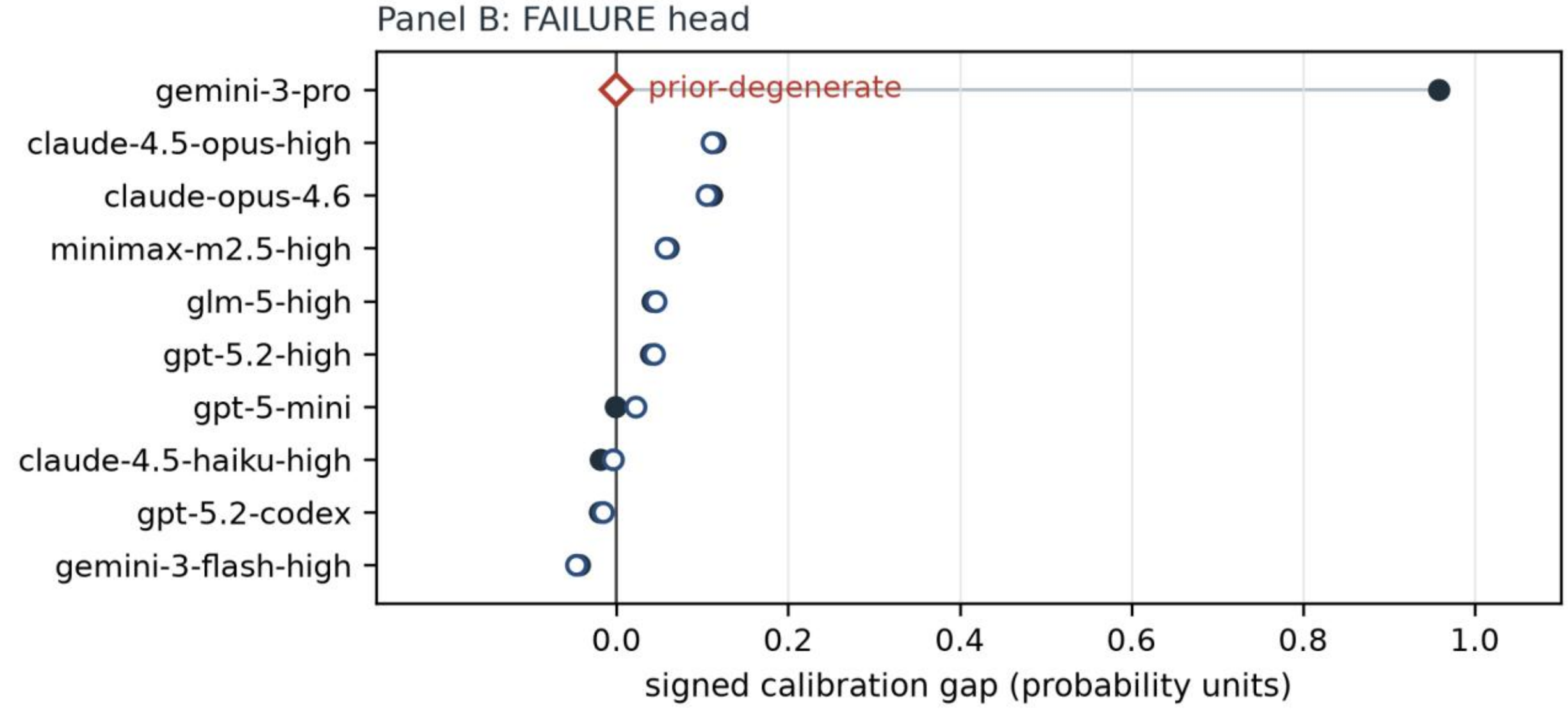

raw signed gap (Phase 0D)
oracle-prior-corrected gap (Phase 0E)
prior-degenerate corrected value (boundary)
frozen primary target cell

Figure 2. Agent-conditional calibration landscape. Each glyph is one held-out model × head cell: filled glyphs show the raw signed calibration gap, open glyphs the same gap after oracle label-prior correction, and panels separate the SUCCESS and FAILURE heads. All values are signed gaps in probability units; the vertical line marks zero. Values come from the frozen Phase 0D audit and Phase 0E decomposition; nothing was recomputed. Cells marked prior-degenerate (gemini-3-pro, both heads) have a held-out success prior of 0 or 1, so the oracle odds ratio is unbounded and the corrected probabilities sit on the boundary; these markers are not comparable gaps. The two frozen primary cells are highlighted (gpt-5-mini/SUCCESS; claude-opus-4.6/FAILURE). The figure shows that prior correction changes the landscape without eliminating all target-conditional residuals; it does not establish same-predictor pairwise heterogeneity, rank agents by quality, or demonstrate a mechanism.

### 4.2 Prior-Shift Decomposition

Correcting for each held-out model's label prior reduces the pooled gaps from 0.0194 to 0.0170 on the SUCCESS head and from 0.1063 to 0.0421 on the FAILURE head but does not remove the largest target-level errors. After the correction, three non-degenerate cells maintained an absolute corrected gap of at least 0.10: gpt-5-mini/SUCCESS at 0.1423, claude-4.5-opus-high/FAILURE at 0.1106, and claude-opus-4.6/FAILURE at 0.1053. The corrected-gap range across the nine non-degenerate models remains 0.1650 on the SUCCESS head (interval [0.1146, 0.2222]) and 0.1579 on the FAILURE head (interval [0.1178, 0.2805]) against the raw ranges of 0.2132 and 0.1584. Therefore, substantial residual cross-model spread remains after oracle correction, and the correction absorbs only part of the observed spread (Figure 2).

This result can be attributed to the two boundary notes. The gemini-3-pro fold is prior-degenerate: its success prior reaches 1.0 on its test folds, the oracle odds ratio is unbounded, and its corrected values sit on the probability boundary (0.0); the gate excludes it, and its markers are not comparable gaps. The correct reading of this stage is the one frozen in advance: the target outcome prior explains only part of the observed calibration variation. The correction does not show that the prior shift is not a cause, and no causal independence is claimed.

### 4.3 Same-Predictor Control

Under the leave-two-agents-out control, where one shared predictor and calibrator score both members of every pair, the broad pairwise heterogeneity hypothesis was not supported. On the SUCCESS head, the median pairwise corrected gap difference is 0.0180 with interval [0.0169, 0.0301] over 45 eligible pairs; on the FAILURE head, it is 0.0385 with interval [0.0318, 0.0640] over 35 eligible pairs. Both values fall below the pre-registered 0.05 median threshold, and the two intervals differ at their lower bound: the SUCCESS-head lower bound of 0.0169 lies below the 0.03 requirement, whereas the FAILURE-head lower bound of 0.0318 lies above it. Each head therefore fails at least one required condition of the frozen heterogeneity rule, and both per-head records are negative, so the frozen heterogeneity gate is recorded as not satisfied. The stage result is recorded as a partial pass: the broad heterogeneity gate is negative, while the target-specific persistence scan is positive for two cells.

The control also clarifies what the negative result does and does not mean. It does not show that the shared predictor is uniformly calibrated across agents; the sustained magnitude in the two cells reported next is evidence against uniformity. The gate shows that the preregistered criterion did not detect broad pairwise heterogeneity when the predictor was held fixed.

## 4.4 Target-Specific Persistence

Two frozen target/head cells retained their calibration-transfer error under every control in the persistence battery. For gpt-5-mini/SUCCESS, the median signed corrected gap is 0.1377 across nine eligible partner-fold occurrences (one occurrence per partner fold, nine of nine in scope), all nine with the same sign; the task-cluster bootstrap interval is [0.1168, 0.1594]; the partner-fold jackknife leaves the signed median unchanged in sign with a minimum absolute value of 0.1362; and the two deterministic task halves give medians of 0.1177 and 0.1489. For claude-opus-4.6/FAILURE, the median is 0.1107 across eight eligible occurrences (nine folds in scope, one occurrence excluded for insufficient decisions), all eight with the same sign; the interval is [0.0510, 0.1486]; the jackknife minimum absolute median is 0.1104; and the task halves give 0.0869 and 0.1135. Both cells satisfied every pre-registered criterion, and the frozen stage result was a strong target-specific signal (Table 2 and Figure 3).

Two qualifications agree with this result. Persistence is strong according to the pre-registered target-specific gate; it is not a statement that agents, in general, are miscalibrated because the same-predictor heterogeneity gate of Section 4.3 is false. The designated sensitivity cell minimax-m2.5-high/FAILURE does not pass (median 0.0785, below the 0.08 magnitude criterion); it is reported as a sensitivity reading and is not promoted to primary evidence.

The pre-registered criteria behind the two verdicts are worth stating in full because they define what "persistent" means here. Each cell had to retain at least seven eligible partner-fold occurrences; reach a median absolute corrected gap of at least 0.08; keep the same sign in at least 80 percent of occurrences; produce a task-cluster bootstrap interval that excludes zero and retains the baseline sign; preserve sign in every leave-one-partner-out jackknife refit with a minimum absolute median of at least 0.06; and keep the baseline sign in both deterministic task halves with at least five eligible folds each. Both primary cells passed all six requirements on the frozen values, with signs preserved in 9 of 9 and 8 of 8 occurrences, respectively, and in both halves, and the jackknife minimum absolute median above the 0.06 floor in both cells.

| | | |
|---|---|---|
| Agent (model label) | gpt-5-mini | claude-opus-4.6 |
| Prediction head | success | failure |
| Eligible same-predictor occurrences | 9 | 8 |
| Partner folds in scope | 9 | 9 |
| Excluded below decision minimum | 0 | 1 |
| Excluded (prior-degenerate) | 0 | 0 |
| Median corrected gap | 0.1377 | 0.1107 |
| Same-sign fraction | 1.0 (9/9) | 1.0 (8/8) |
| Bootstrap 95% CI (task-cluster) | [0.1168, 0.1594] | [0.0510, 0.1486] |
| Jackknife min abs median | 0.1362 | 0.1104 |
| Task-half A median | 0.1177 | 0.0869 |

| Task-half B median | 0.1489 | 0.1135 |
|---|---|---|
| Threshold robustness | 4/4 evaluable thresholds robust (0.900/0.925/0.950/0.975) | 3/3 evaluable thresholds robust (0.900/0.925/0.950); 0.975 not evaluable (denominator) |
| Final target status | Robust | Robust |

Footnotes. (1) All gap values are oracle-prior-corrected signed gaps (predicted confidence minus the realized outcome rate at the early stop point). (2) Interval-source rule: The persistence columns (bootstrap CI, jackknife, task halves) come from Phase 1A-D; the threshold column comes from Phase 1B, whose bootstrap execution is distinct, so the two sets of intervals are kept separate. (3) 0.975 is not applicable for claude-opus-4.6 / FAILURE because no occurrence clears the frozen decision count minimum at that threshold. (4) Unit of analysis: One eligible partner-fold occurrence of the target cell. (5) Final target status 'Robust' transcribes the frozen Phase 1A-D result STRONG_TARGET_SPECIFIC_SIGNAL; 'not evaluable' transcribes the frozen threshold-sweep state.

Table 2. Primary SWE-bench Verified target results for the two frozen target/head cells. Values are oracle-prior-corrected signed calibration gaps from the Phase 1A-D persistence battery (eligible occurrences, median, same-sign fraction, task-cluster bootstrap 95% CI, partner-fold jackknife minimum, task-half medians), with the threshold-robustness column from the distinct Phase 1B bootstrap execution. The table supports the statement that the errors of both cells persist across source-agent cohorts, task resampling, and stopping thresholds within this frozen setting. It does not establish that calibration failure is broad across agents, and it does not generalize beyond SWE-bench Verified.

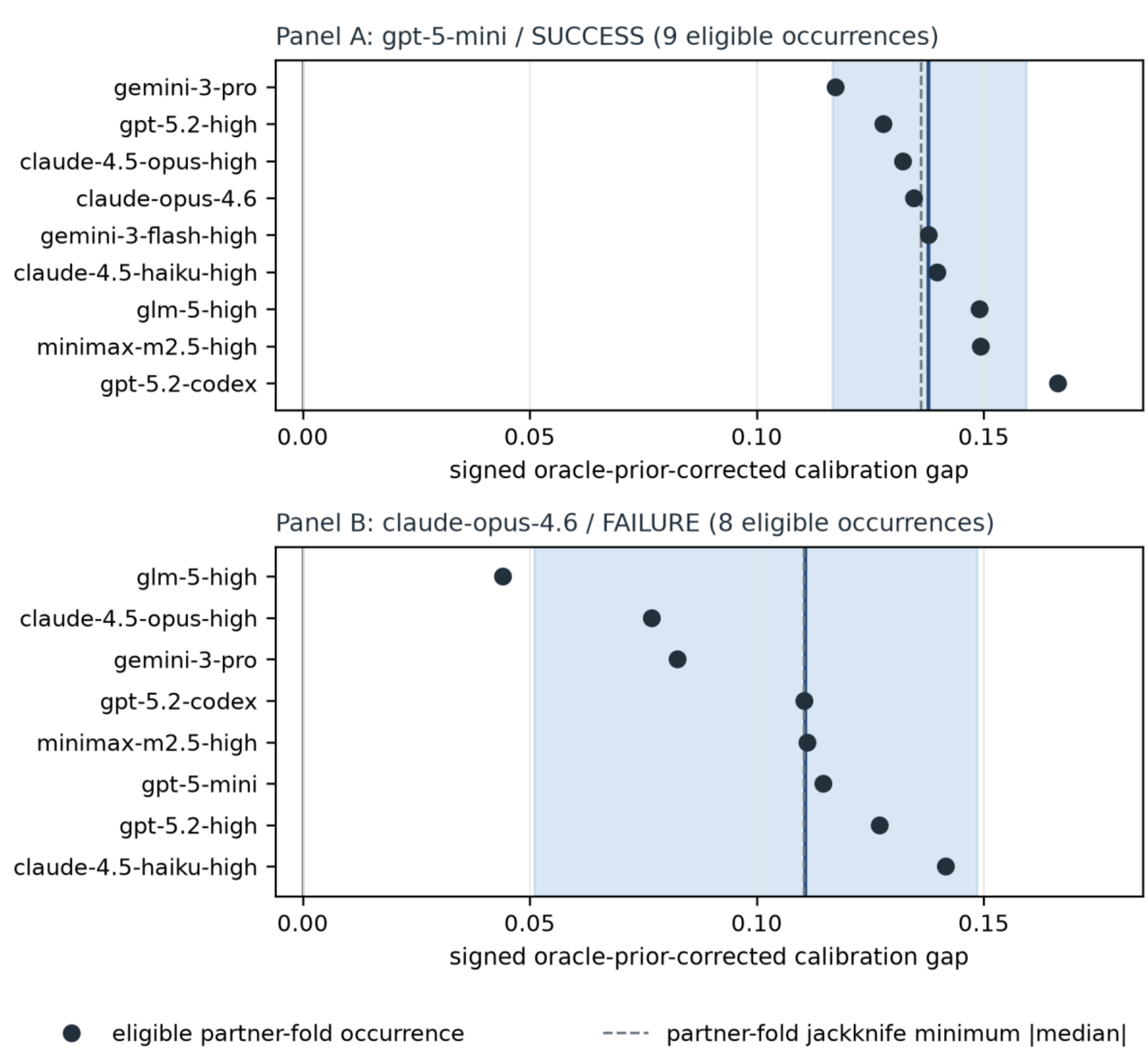

Figure 3. Target-specific persistence of two SWE-bench Verified calibration-transfer errors under a shared predictor. One glyph per eligible partner-fold occurrence. The y-axis lists the partner held-out agent, and the x-axis shows the signed oracle-prior-corrected calibration gap, with a vertical line at zero. Panel A: gpt-5-mini/SUCCESS, 9 eligible occurrences. Panel B: claude-opus-4.6/FAILURE, 8 eligible occurrences; one further occurrence fails the frozen eligibility rule and is excluded from the plotted gap values. Overlays mark the panel median, its task-cluster bootstrap 95% CI, the partner-fold jackknife minimum, and the two task-half medians; they belong to the panel median, and no CI is drawn for an individual occurrence. Values are from the frozen Phase 1A shared-predictor control and Phase 1A-D persistence validation. The figure establishes target-specific persistence within this setting; it does not establish broad pairwise heterogeneity, and it does not generalize beyond SWE-bench Verified.

## 4.5 Threshold Robustness

Re-applying the unchanged policy across the frozen threshold grid shows that neither target error is tied to the single reference threshold. The 0.950 anchor reproduces all 44,901 frozen decision rows exactly (maximum absolute decision-score difference of $1.11 \times 10^{-16}$ against a declared tolerance of $1 \times 10^{-9}$); therefore, the sweep starts from a verified reproduction. For gpt-5-mini/SUCCESS all four grid points pass: 0.900 gives 0.1125 with interval [0.0954, 0.1268]; 0.925 gives 0.1200 [0.1019, 0.1365]; 0.950 gives 0.1377 [0.1175, 0.1599]; and 0.975 gives 0.1698 [0.1274, 0.2144]. For claude-opus-4.6/FAILURE three of four grid points are robust: 0.900 gives 0.1374 [0.0895, 0.1680], 0.925 gives 0.1160 [0.0838, 0.1707], and 0.950 gives 0.1107 [0.0510, 0.1462]. At 0.975, the target has no eligible occurrence at all (0 eligible folds; all 8 occurrences in scope fall below the frozen decision minimum, with per-occurrence head decisions between 8 and 16), so the point is not evaluable and is reported as such; no value is imputed and the point is not evidence of absence. The frozen stage result was a strong pass (Figure 4, panel A).

The per-occurrence decision counts show the number of samples carried by each grid point. For gpt-5-mini/SUCCESS, the eligible occurrences hold between 282 and 313 head decisions at 0.900, between 248 and 298 at 0.925, between 177 and 269 at 0.950, and between 26 and 142 at 0.975; for claude-opus-4.6/FAILURE, the counts run between 60 and 71 at 0.900, 48 and 58 at 0.925, and 32 and 40 at 0.950. The high-threshold end of the grid is therefore thin for the failure target, even where it remains evaluable, which is why the sweep's verdict is reported as three evaluable points rather than four.

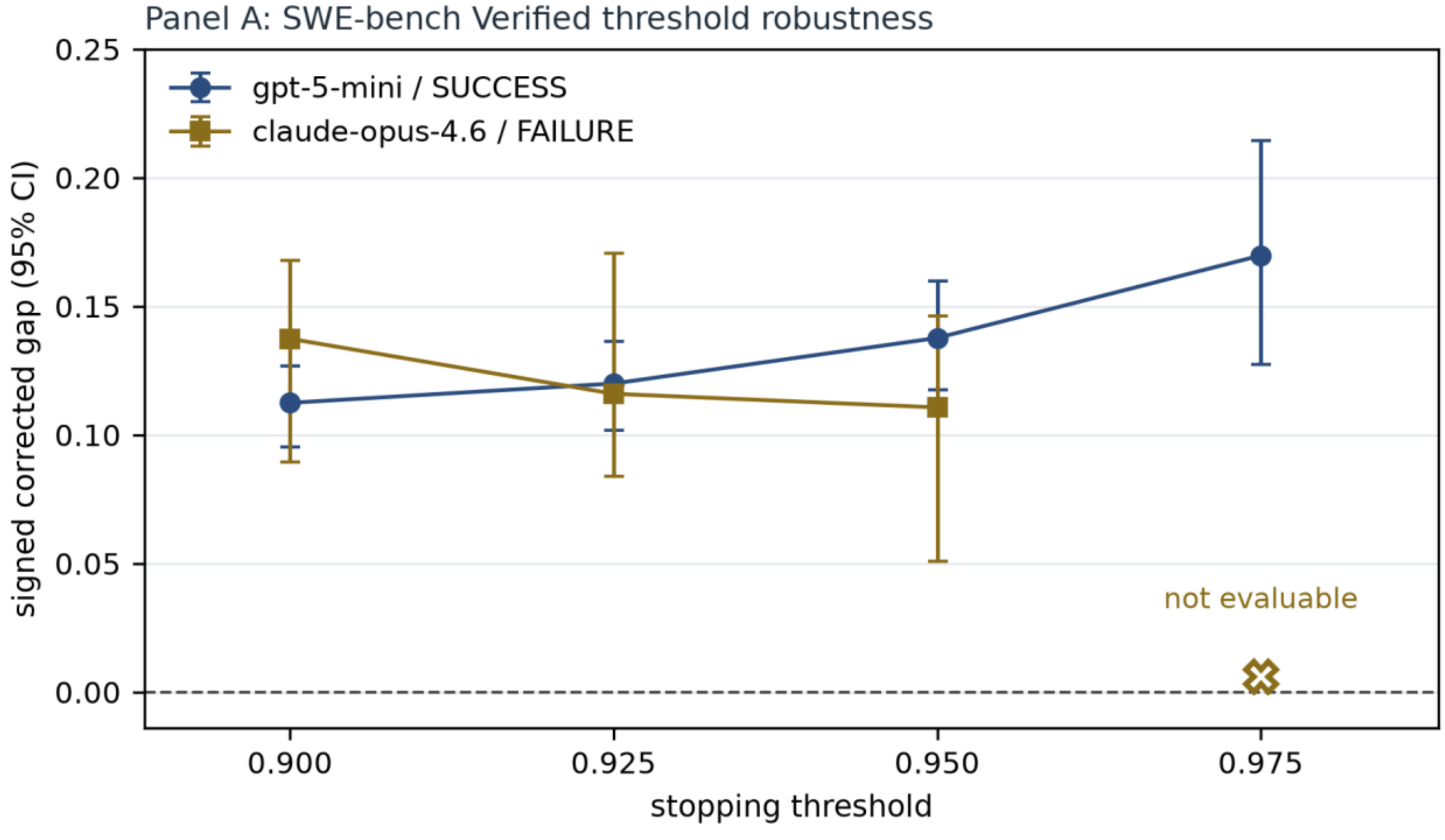

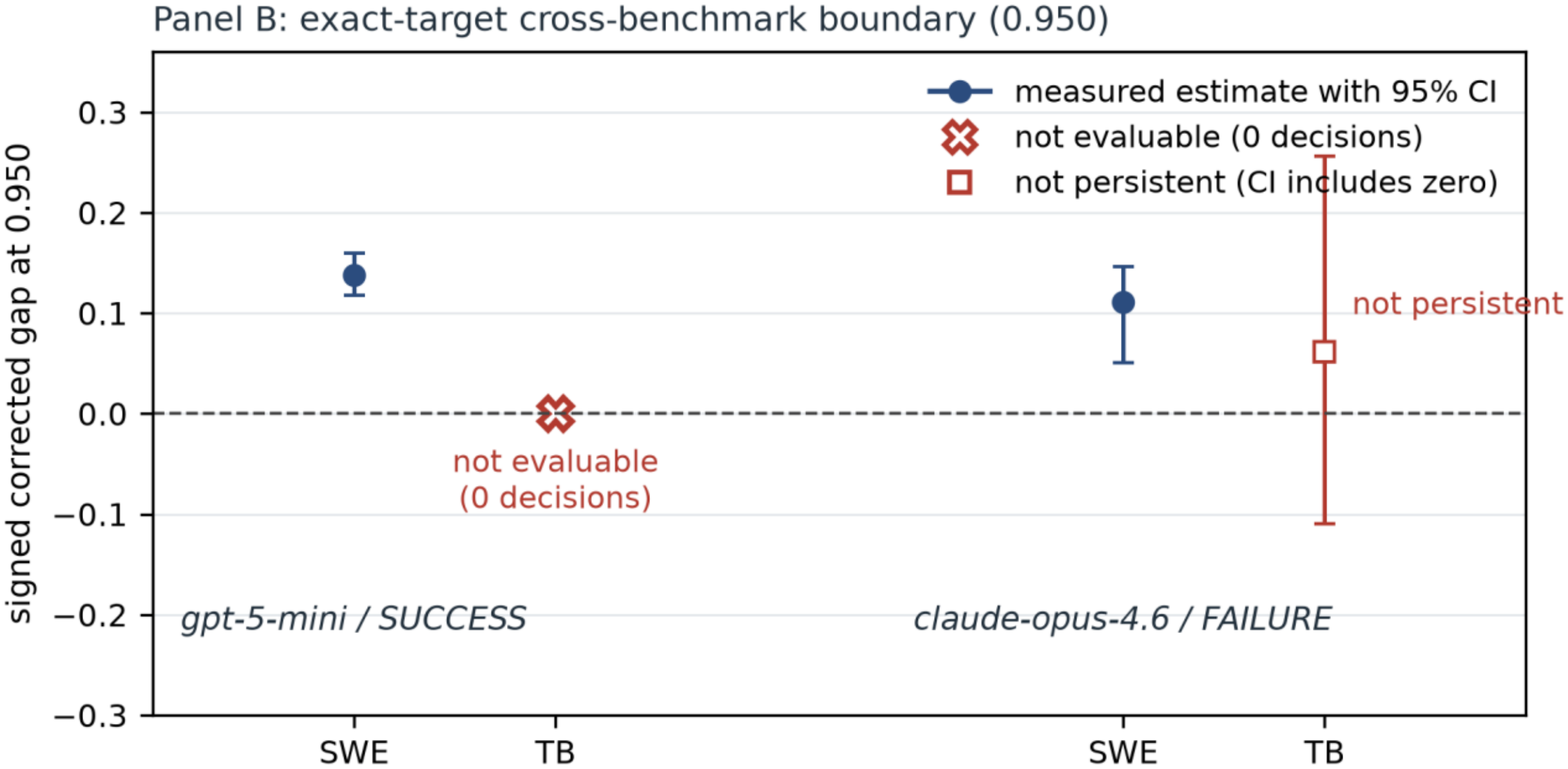

Figure 4. Threshold robustness and cross-benchmark boundary. Panel A: Signed oracle-prior-corrected calibration gap against the stopping threshold (0.900, 0.925, 0.950, 0.975) for the two frozen SWE targets, with frozen task-cluster bootstrap 95% CIs. The claude-opus-4.6/FAILURE series is not evaluable at 0.975 because no occurrence clears the decision-count minimum; that state is drawn as a not-evaluable marker rather than interpolated or set to zero. Panel B: Exact-target cross-benchmark comparison at 0.950 for the same two targets on SWE-bench Verified versus TerminalBench under the fixed terminus-2 scaffold. For the SWE-bench points, the plotted estimate is the frozen target-level median corrected gap at 0.950 with its task-cluster bootstrap interval; for TerminalBench, the plotted value is the identity-matched held-out target-model estimate at 0.950. Marker classes distinguish measured estimates with confidence intervals, non-evaluable cells, and non-persistent cells. The gpt-5-mini/SUCCESS target produced 0 TerminalBench decisions,

classifying it as indeterminate: an absence of measurement rather than a zero gap or a collapse. The claude-opus-4.6/FAILURE target produced 45 decisions with an interval that includes zero, so it is recorded as not persistent under the frozen criterion. Values are from the frozen Phase 1B, 2B, 2B-D, and 2E artifacts. Panel A shows that the target-specific error is not an artifact of a single stopping threshold. Panel B is a boundary result: it does not establish cross-benchmark replication, and it does not establish that the benchmark or the environment caused the discrepancy.

## 4.6 TerminalBench Cross-Benchmark Boundary Test

The second benchmark test returns negative and partly indeterminate evidence. Phase 2B completed all 29 leave-one-model-out folds with zero failures inside the fixed `terminus-2` scaffold, covering 15,889 trajectories across 89 tasks, of which 3,784 (coverage: 0.2382) produced a decision at the primary threshold of 0.950. The SUCCESS head has a single eligible target, with a median absolute corrected gap of 0.0154; the FAILURE head has 27 eligible targets, with a median absolute corrected gap of 0.0102, a maximum of 0.1062, and a range of 0.1831. Exactly one target/head cell satisfies the large-gap rule: gemini-3-pro-preview@gemini/FAILURE, with 23 decisions, precision 1.0, corrected mean score 0.8938, signed corrected gap −0.1062, and interval [−0.1144, −0.0986]. The gate requires at least three such models from more than one provider family; therefore, the frozen-stage result is a negative target-specific signal gate and no cross-benchmark signal.

The threshold and denominator diagnostic asks whether the primary negative result persists across nearby thresholds. The signal gate is satisfied only at 0.900, but not at 0.925 or 0.950; consequently, the preregistered diagnostic does not support cross-benchmark replication. The diagnostic also exposes the denominator that limits the SUCCESS head: the head has 11 eligible targets at 0.900, 3 at 0.925, and 1 at 0.950, and the frozen adequacy rule requires at least 8 eligible targets. The diagnostic is reported here as diagnostic; it does not replace the failed primary gate, and its 0.900 and 0.925 per-target values are kept in the appendix.

One concise reference to secondary descriptions is needed for transparency. Under the frozen secondary-threshold classification, the failure target is recorded at 0.900 and 0.925 with the frozen label `COLLAPSED_ON_TERMINALBENCH`, which denotes a small TerminalBench gap whose interval includes zero, and as unclassified at 0.950; the frozen protocol selects no threshold as best, builds no gate from those rows, and keeps them in the appendix. They are not used here to soften or strengthen the primary 0.950 boundary result.

The exact-identity audit pins both primary SWE targets to their TerminalBench counterparts from literal dataset metadata (gpt-5-mini maps to gpt-5-mini@openai; claude-opus-4.6 maps to claude-opus-4-6@anthropic) and then applies the frozen four-clause status rule at 0.950. For gpt-5-mini/SUCCESS, TerminalBench produced 0 decisions at 0.900, 0.925, and 0.950; thus, the cross-benchmark status is INDETERMINATE: nothing was measured for that cell, and the reading is not a collapse. For claude-opus-4.6/FAILURE, TerminalBench produced 45 decisions over 14 tasks with an empirical precision of 0.8, a corrected mean score of 0.8620, and a signed corrected gap of +0.0620 with an interval [−0.1094, 0.2563]. The interval includes zero; thus, the cell does not satisfy the pre-registered persistence criterion and is treated as nonpersistent. Under the exact 0.950 status clauses, the point falls in a region the four clauses do not exhaust (UNCLASSIFIED_BY_SPEC, recorded as a deviation), while its sign agrees with the SWE gap. The overall frozen result of this stage is insufficient cross-benchmark support (Table 3; Figure 4, panel B).

**Section A. Fixed-scaffold (terminus-2) replication summary.**

| Item | Value |
|---|---|
| Eligible models under terminus-2 | 29 |
| Completed leave-one-model-out folds | 29/29 |
| Usable trajectories | 15,889 |
| Decided at 0.950 | 3,784 |
| Decision coverage at 0.950 | 0.2382 |
| Eligible targets at 0.950 (SUCCESS head) | 1 (median \|gap\| = 0.0154, max = 0.0154) |
| Eligible targets at 0.950 (FAILURE head) | 27 (median \|gap\| = 0.0102, max = 0.1062, range = 0.1831) |
| Robust large-gap rows | 1 (gemini-3-pro-preview@gemini/FAILURE; gap -0.1062, CI [-0.1144, -0.0986]) |
| Primary replication gate (Phase 2B) | Not supported |
| Threshold diagnostic gate (Phase 2B-D) | Not supported |

**Section B. Exact-target boundary at 0.950.**

| Target / head | Benchmark | Model label (identity) | Decisions | Gap at 0.950 | Status |
|---|---|---|---|---|---|
| gpt-5-mini / SUCCESS | SWE-bench | gpt-5-mini | 1,860 | 0.1377 | Persistent on SWE-bench |
| gpt-5-mini / SUCCESS | TerminalBench | gpt-5-mini@openai | 0 | NA | Indeterminate (no measurement) |
| claude-opus-4.6 / FAILURE | SWE-bench | claude-opus-4.6 | 289 | 0.1107 | Persistent on SWE-bench |
| claude-opus-4.6 / FAILURE | TerminalBench | claude-opus-4-6@anthropic | 45 | +0.0620 | Not persistent; unclassified by the frozen 0.950 rule |

Footnotes. (1) Section B gaps are oracle-prior-corrected signed gaps at the frozen 0.950 threshold. (2) gpt-5-mini / SUCCESS produces 0 TerminalBench decisions; its gap is NA (absence of measurement), never 0, and its status is INDETERMINATE rather than any form of collapse. (3) claude-opus-4.6 / FAILURE produced 45 TerminalBench decisions; its interval includes zero, so it is recorded as not persistent under the frozen criterion, with the 0.950 point lying in an intermediate region not exhausted by the pre-registered classification categories. (4) Mandatory statement: Failure to replicate does not establish that the benchmark/environment causes the discrepancy. (5) Exact frozen status strings: Section A gates are NO_CROSS_BENCHMARK_SIGNAL (with TERMINAL_TARGET_SPECIFIC_SIGNAL = FALSE) and NO_CROSS_BENCHMARK_REPLICATION (THRESHOLD_SIGNAL 1 of 3 at 0.900); Section B statuses are PERSISTENT_ON_SWE_BENCH, INDETERMINATE, and NOT_PERSISTENT;UNCLASSIFIED_BY_SPEC_AT_0.950. All values are transcribed from the frozen Phase 2B, Phase 2B-D, and Phase 2E records listed in the supplement.

Table 3. TerminalBench boundary results under the fixed terminus-2 scaffold. Section A summarizes the fixed-scaffold replication (29 eligible models, 29/29 completed leave-one-model-out folds, 15,889 usable trajectories, 3,784 decided at 0.950, decision coverage 0.2382) and reports the two frozen gates. Section B compares the two SWE primary target cells with their identity-matched TerminalBench counterparts at 0.950: gpt-5-mini/SUCCESS produced 0 TerminalBench decisions, so its TerminalBench status is indeterminate and its gap is NA (absence of measurement, never zero); claude-opus-4.6/FAILURE produced 45 decisions with an interval that includes zero and is recorded as not persistent under the frozen criterion. Exact frozen status strings are transcribed in the footnotes. The table supports the statement that the SWE target-specific pattern is not established as benchmark-general under this scaffold and criterion; failure to replicate does not establish that the benchmark or environment causes the discrepancy, and the table does not show that the SWE effect disappears.

## *5. Discussion*

### 5.1 Target-Specific Rather Than Broad Transfer Failure

The central contrast of the study is between a negative population-level result and a positive localized one. Under a shared predictor, the median pairwise corrected-gap difference was 0.0180 on the SUCCESS head and 0.0385 on the FAILURE head, and neither head satisfied the pre-registered heterogeneity criterion. At the same time, two target/head cells carried median-corrected gaps of 0.1377 and 0.1107 that maintained their sign across cohort changes, task resampling, task halves, partner jackknife, and thresholds. Small pairwise dispersion and two persistent large cells can coexist, and when they do, target/head-level analysis reveals information that population averages can obscure.

This reading has firm boundaries. It does not establish that the two cells are representative of a class of agents, and it does not support the claim that unseen-agent transfer is generally unreliable; the negative broad gate is part of the result, not an obstacle to it. It also does not show that the shared predictor is uniformly calibrated because the two persistent cells themselves depart from uniformity. The frozen record supports the narrower statement that the calibration-transfer error in this setting is concentrated rather than broad.

It is useful to separate the layers of this statement. The observed facts are the frozen-gate values and their stability across the six controls. The interpretation drawn here is that the useful unit of analysis for reliability is the target/head cell rather than the population of agents. Unresolved alternatives remain, and the frozen record does not choose between them: the two cells differ in their decision denominators and in the composition of the tasks that reach a stop; the predictor's feature-level behavior on those cells was not analyzed; and the prior correction is an oracle reweighting rather than an intervention. The supported scope is accordingly the frozen SWE-bench Verified environment, the two named cells, and the frozen control battery.

### 5.2 Why Aggregate Calibration Can Hide Localized Errors

The pooled calibration numbers appear benign relative to the worst cells. After oracle prior correction, the pooled gaps are 0.0170 on the SUCCESS head and 0.0421 on the FAILURE head, while three non-degenerate cells sit at 0.1423, 0.1106, and 0.1053. The pooled values are aggregates over many cells with smaller residual gaps, and on the FAILURE head, they also shift substantially with the label prior (from 0.1063 raw to 0.0421 corrected); however, the same correction leaves the largest target-level errors essentially in place. A report that showed only the pooled gap or only the raw gap would have hidden both the concentration and role of the prior.

This is an observation of one predictor's audit, not a new claim about aggregation. The literature has already established that pooled and global calibration comparisons can mislead: rankings can reverse under accuracy control [12], a pooled monitor average can conceal fragility on a subset [13], and trajectory averaging can hide stage-specific overconfidence [14]. The contribution here is the application of this principle to transferred early outcome confidence and the observation that, in this frozen setting, the hidden object is a small set of target/head cells.

### 5.3 Interpreting the TerminalBench Non-Replication

TerminalBench analysis showed that the SWE target-specific pattern was not established under the frozen TerminalBench setting. Phase 2B finds a single large-gap model (one provider family) against a gate that requires at least three across more than one family; Phase 2B-D satisfies the diagnostic signal gate at only one of three thresholds; and at exact model identity, the SUCCESS target produces no decisions at all while the FAILURE target produces 45 decisions over 14 tasks with an interval that includes zero. Both readings are constraints on what can be concluded, and the SUCCESS-head reading is an absence of measurement rather than a measured change.

This result does not support causal language. No design component manipulates the benchmark or environment; therefore, the discrepancy cannot be attributed to the benchmark; scaffold identity is fixed rather than varied, so it cannot be attributed to the scaffold; and a failed replication criterion is not a disconfirmation of the SWE finding. A negative boundary result of this kind is informative precisely because calibration conclusions are known to be unstable across models, benchmarks, and stages [12,9,8,4], and because the second benchmark's success-head denominator is too small to answer the question the SWE analysis could answer. The honest summary is that the phenotype is established within one frozen evaluation environment and remains unestablished.

The layers of this boundary result can be stated compactly. The observed facts are the fixed-scaffold gate outcomes, exact-identity measurements, and identity of one large-gap TerminalBench target. The interpretation is that the SWE pattern is not established as benchmark-general under the freezing criterion. Unresolved alternatives include the possibility that the second benchmark's coverage (scaffold, task set, decision denominators) is simply different in ways this design does not isolate, that the two benchmarks differ in the behavior of the trajectories they collect, that the compared models differ in training distribution, and that model/scaffold interactions contribute in ways a single fixed scaffold cannot reveal. The experiment could not separate the benchmark from the environment it was embedded in, and it was not designed to. The supported scope is a constraint on how the SWE phenotype may be read and nothing more.

## 5.4 Implications for Early-Stopping Evaluation

The practical implications of this audit concern measurement practice. First, the calibration of an early outcome predictor should be reported per target/head cell alongside any aggregate figure because a policy that behaves well on average can still commit at high confidence on a target where its confidence is wrong. Second, when several agents are compared, a shared-predictor control helps distinguish target-related differences from variation introduced by retraining, and is inexpensive to add once the pair design exists. Third, aggregate transfer metrics should not be read as evidence that confidence thresholds transfer; in this record, the aggregate view appeared benign while two cells remained miscalibrated. Fourth, before aggressive early stopping is deployed in a new agent system, the calibration of the decisions it would take on that system should be checked directly.

These reporting and testing recommendations are grounded in the observed record. No mitigation method was evaluated: the frozen record contained no recalibration, abstention, or policy-repair experiment, and nothing here should be read as a validating one.

## 5.5 Limitations

First, strong positive evidence is confined to one benchmark. All continuous positive results come from a single frozen SWE-bench Verified corpus, and the second benchmark functions as a boundary test rather than a second confirmation.

Second, the primary target-specific finding rests on two frozen agent/head cells, gpt-5-mini/SUCCESS and claude-opus-4.6/FAILURE. The sensitivity cell minimax-m2.5-high/FAILURE did not satisfy the magnitude criterion (0.0785 vs. 0.08) and did not provide primary evidence.

Third, no causal mechanism was established. Every control was observational or oracle reweighting of the same data; nothing intervened in the predictor, agent, or environment.

Fourth, the TerminalBench SUCCESS head is denominator-limited: one eligible target at 0.950 (11 at 0.900 and 3 at 0.925), which is below the frozen adequacy requirement for that head, and the FAILURE target's 0.975 threshold point is not evaluable in the SWE threshold sweep.

Fifth, the TerminalBench public trajectory source is community-compiled, and the trial-level identity overlap between the community-compiled TerminalBench dataset and the historical TerminalBench corpus/provenance used by the upstream EarlyEval release could not be verified from the frozen record; the analysis treats the second benchmark as a boundary setting rather than as an independent sample.

Sixth, no mitigation method was evaluated; therefore, the study cannot say which recalibration, abstention, or policy-repair strategy would help.

Seventh, the prediction framework is closely related to prior work, and much of it is inherited: the reference-free dual-head predictor, calibrated threshold policy, and leave-one-agent-out protocol are reused from the early outcome-prediction line [1], and the shared-predictor control and persistence battery are the parts added here.

Two further limitations were recorded for completeness. The initial audit's FAILURE-head qualification is a degenerate-fold artifact that disappears when the prior-degenerate fold is excluded (range 1.0, falling to 0.1475); therefore, the initial qualification should be read through the SUCCESS head. One TerminalBench status at 0.950 falls outside the pre-registered four-clause classification and is recorded as a deviation rather than a reinterpretation.

## *6. Conclusion*

This study audited the calibration of a frozen trajectory-based early outcome predictor when it is transferred to unseen agents. The broad hypothesis that same-predictor transfer error is heterogeneous across agent pairs was not supported: median pairwise corrected-gap differences were small, and the pre-registered heterogeneity criterion was not met on either head. Within the same frozen SWE-bench Verified setting, however, two specific target/head cells, gpt-5-mini/SUCCESS and claude-opus-4.6/FAILURE, showed calibration transfer error of persistent sign and magnitude across changing source-agent cohorts, task resampling, partner jackknife, task halves, and stopping thresholds, and survived oracle prior correction. A fixed-scaffold TerminalBench boundary test did not establish a pattern beyond the first benchmark; the success target produced no decisions, and the failure target did not satisfy the pre-registered persistence criterion. The conservative implication is procedural: the calibration of early outcome predictors should be audited at the level of the target system before confidence thresholds are treated as transferable because aggregate transfer metrics can appear benign while specific target/head cells remain miscalibrated.

## *Conflict of Interest*

The author declares that he has no known competing financial interests or personal relationships that could have influenced the work reported in this paper.

### *Data Availability*

This study re-analyzed two third-party public trajectory datasets and did not redistribute either of them. The primary corpus is the frozen SWE-bench Verified agent-trajectory collection (`tarsur385/swebench-verified-trajectories`, dataset revision `773748a7c1222e8a642a7059821498e14293562a`, MIT); the second corpus is the community-compiled TerminalBench trajectory dataset (`yoonholee/terminalbench-trajectories`, dataset revision `04e8940f5b6736a7ce8d22224fe2f2af74163ed2`, Apache-2.0). Both are consumed in the revisions recorded in Table 1 and in the frozen manifests, and users should obtain both corpora from their original sources.

The derived analysis artifacts of this study–the per-phase frozen result tables, per-cell calibration summaries, bootstrap outputs, gate records, and the plot and table source data–are publicly available, with manifests and per-file SHA-256 checksums, in the public repository at https://github.com/caoyanze426-crypto/early-outcome-calibration-audit. The author-produced derived artifacts are released under CC BY 4.0 (`LICENSE-DATA`) and the analysis code under the MIT License (`LICENSE`); third-party material retains its original license and is not redistributed. The artifact inventory and checksums that accompany the frozen phases are available to reviewers on request and are the records on which every number in this paper is based.

### *Code Availability*

The analysis code developed for this study is publicly available in the repository at https://github.com/caoyanze426-crypto/early-outcome-calibration-audit, under the MIT License, together with the frozen gate definitions it implements. Third-party components maintain their original provenance and licensing, as recorded in the repository's `THIRD_PARTY_NOTICES.md`. The audited design is built on the public early-outcome-prediction release [1]: its stopping-policy module is used unmodified at the frozen commit `7fd1a8e5b755e1f7ab642bcae77473b53e2bf1d0`, and the upstream component remains under its original upstream licensing and provenance; the audited predictor is an EarlyEval-style dual-head model fitted inside this study's pipeline with the release's hyperparameters and the reference-free feature configuration. The upstream release is code-only and ships no trained predictor; the fitted models audited here were produced by this study and form part of the frozen artifacts.

### *Funding*

This research did not receive any specific grants from funding agencies in the public, commercial, or not-for-profit sectors.

### *Declaration of generative AI and AI-assisted technologies in the manuscript preparation process*

GPT-5.6 Sol was used during the research stage to assist with the literature search. The research direction was led by the human author, and the literature identified with AI assistance was reviewed and confirmed by the author.